\documentclass[letterpaper, 10 pt, conference]{ieeeconf}
\IEEEoverridecommandlockouts
\usepackage[colorlinks=true,allcolors=black,breaklinks]{hyperref}
\usepackage[switch]{lineno}
\usepackage{algorithm}
\usepackage{algorithmic}

\usepackage{geometry}
\usepackage{amsmath}
\usepackage{amssymb}
\usepackage{cleveref}
\usepackage{cite}
\usepackage[usenames, dvipsnames]{color}
\usepackage{graphicx}
\usepackage{siunitx}
\usepackage{lmodern}
\usepackage{subcaption}
\usepackage[font=small]{caption}
\usepackage{balance}

\usepackage{float}

\crefname{figure}{Figure}{Figures}
\crefname{algorithm}{Alg.}{Algorithms}
\crefname{equation}{Equation}{Equations}
\crefname{section}{Section}{Sections}
\crefname{appendix}{Appendix}{Appendix}

\newcommand{\old}[1]{{}}
\newcommand{\bigO}{\mathcal{O}} 
\newcommand{\points}{S}
\newcommand{\point}{p}
\newcommand{\smallFixM}{\emph{uni\_sm}}
\newcommand{\largeFixM}{\emph{uni\_lg}}
\newcommand{\fixN}{\emph{uni\_fix\_n}}

\newcommand{\gmcip}{GMC IP}
\newcommand{\dgmcip}{DGMC IP}
\newcommand{\iteralg}{GMC heuristic}

\title{\LARGE\bf  Multi-Covering a Point Set by $m$ Disks\\ with Minimum Total Area}
\author{Mariem Guitouni$^{1,\dagger}$,
    Chek-Manh Loi$^{2,\dagger}$,
    Sándor Fekete$^2$,
    Michael Perk$^2$,
    Aaron T. Becker$^1$%
    \thanks{$^\dagger$ These authors share lead authorship}
    \thanks{$^1$mguitoun@CougarNet.uh.edu, atbecker@uh.edu; partially supported by NSF IIS-2130793,  ARL  W911NF-23-2-0014, and DAF AFX23D-TCSO1.
    }
\thanks{$^2$loi@ibr.cs.tu-bs.de, s.fekete@tu-bs.de, perk@ibr.cs.tu-bs.de; partially supported by DFG grant FE407/21-1.}}

\begin{document}
\maketitle
\begin{abstract}
    A common robotics sensing problem is to place sensors to robustly monitor a set of assets, where robustness is assured by requiring  asset $p$ to be monitored by at least $\kappa(p)$ sensors.
    Given $n$ assets that must be observed by $m$ sensors, each with a disk-shaped sensing region, where should the sensors be placed to minimize the total area observed?  We provide and analyze a fast heuristic for this problem.
    We then use the heuristic to initialize an exact Integer Programming solution.
    Subsequently, we enforce separation constraints between the sensors by modifying the integer program formulation and by changing the disk candidate set.
\end{abstract}

\section{Introduction}
\begin{figure}[t]
    \centering
    \includegraphics[width=\columnwidth, trim={0 0.5cm 0.5cm 0},clip]{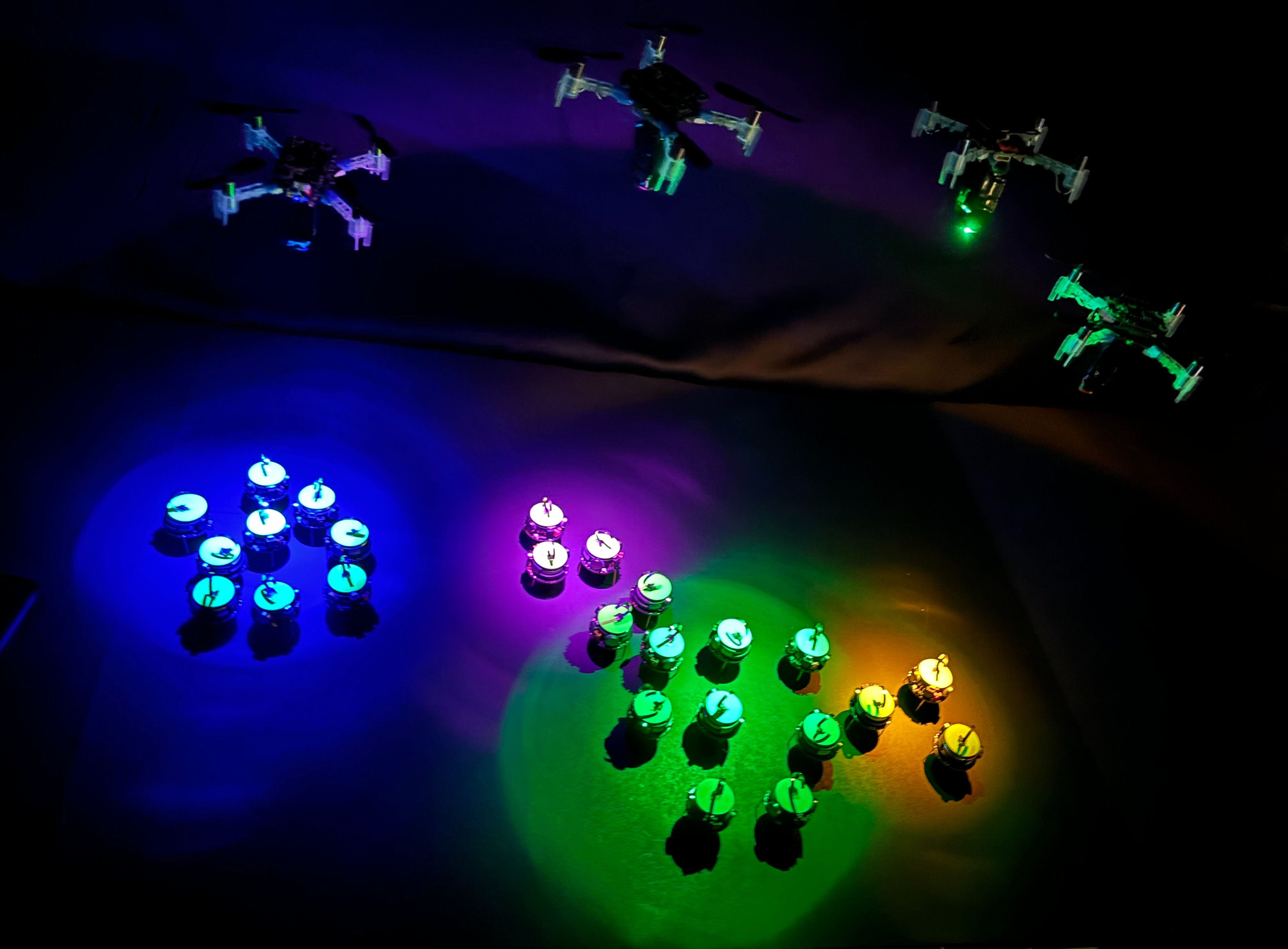}
    \caption{\label{fig:drone}
    Ground-based Kilobot robots, commanded by overhead controllers via infrared communication~\cite{rubenstein2012kilobot}.  
    A LED spotlight underneath each drone represents the communication link.  25 Kilobots are covered by at least one drone, but three high-value robots are multi-covered by two drones. 
    To minimize energy consumption, it is desirable to make the coverage disks as small as possible.
    }
    \vspace{-1em}
\end{figure}

Coordinating different kinds of robotic assets is a natural challenge when it comes to problems of surveillance and guidance.
As shown in ~\cref{fig:drone}, this gives rise to scenarios
in which a finite set of drones with downward communication links must maintain control of 
a finite set of ground assets~\cite{liu2023robust,Boudjit2015}, choosing a set of different altitudes that 
balance safe separation between drones with reliable communication to the ground.
The latter requires sufficient signal strength, so communication areas (and thus energy consumption) depend quadratically on the altitude.
Coverage redundancy -- observing the same target with more than one drone -- is often needed because flying drones are subject to unpredictable failures.
The desired level of redundancy depends on the value of the target.
Similar problems exist in diverse application domains, including wireless sensor networks \cite{abu2011multi,bar2013note,bernardini2024,10611561,10619030}, facility placement \cite{DBLP:conf/compgeom/AltABEFKLMW06,bhowmick2013constant} and pesticide application~\cite{lal2021time,10160265}.

In general, we want to cover a set $\points$ of $n$ assets by placing $m$ disks of minimum total area.
Given a point set $\points$, a number of sensors $m$, and a coverage function $\kappa: \points \to \mathbb{N}$,
the goal is to assign a center $y_i$ and radius $r_i$ for each disk $i \in [1,\dots,m]$ so that each $\point \in \points$ is covered by at least $\kappa(\point)$ disks.
The objective function to minimize is the sum of the areas of the disks $\pi \sum_{i=1}^{m} r_i^2$.
Alt~et~al.~\cite{DBLP:conf/compgeom/AltABEFKLMW06} studied a version for which $\kappa(\point)=1$.
We call our generalization the \emph{general multi-coverage} (GMC) variant.
In our real-world application, drones must have sufficient separation to avoid collisions and occlusions.
Given some distance $\ell$, the \emph{dispersive multi-coverage} problem (DGMC) asks for a GMC with $\lVert y_i -y_j \rVert \geq \ell$ for all $i\neq j$ and $i,j\in [1,\dots,m]$.

\begin{figure}[ht!]
    \centering
    \includegraphics[width=.95\columnwidth, trim={0 0.5cm 0 0},clip]{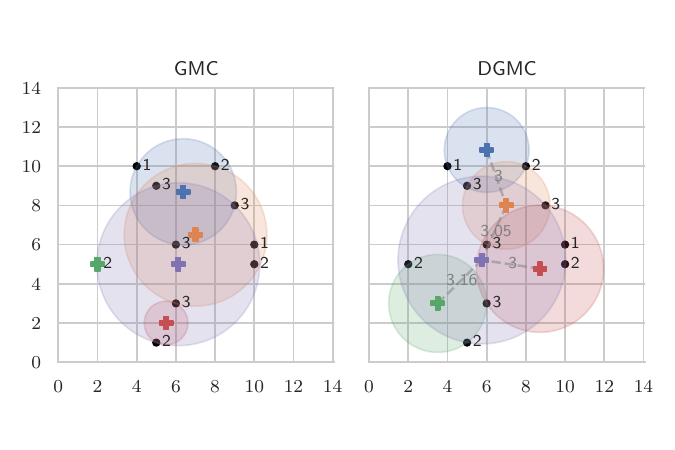}
    \caption{\label{fig:optimalsolutionexample}
        Optimal (minimum total area) solutions for the GMC and DGMC with $n=10$, $m=5$, and $\kappa$ between $1$ and $3$.
        (Left) Without separation constraints, the sum of the areas is equal to \SI{122.93}{\metre\squared}.
        Note that there is a radius $0$ disk on the leftmost point.
        (Right) Enforcing a minimum distance of $\ell=3$ yields an optimal solution with a total area of \SI{140.66}{\metre\squared}.
        The gray lines indicate the distance between the disks.
    }
\end{figure}

The example in \cref{fig:optimalsolutionexample} shows the optimal solution 
for a problem with $n=10$ assets, $m=5$ sensors and $\kappa$ between 1 and 3 as well as an optimal solution for the DGMC with $\ell=3$.
Each point $\point \in \points$ is annotated by its coverage requirement $\kappa(\point)$.

The remainder of this paper is organized as follows.
In \cref{sec:RelatedWork}, we review related work. 
\cref{sec:LB} presents two approaches to solve the GMC problem: an iterative heuristic and Integer Programming (IP).
\cref{sec:UB} introduces an IP approach for computing provably optimal solutions for DGMC instances.
In \cref{sec:Results} we provide a comparison between the iterative heuristic and the corresponding optimal solutions of the GMC.
Finally, we compare different strategies for the DGMC.

\section{Related Work}\label{sec:RelatedWork}

A related, but simpler, problem explored in previous works uses given disk centers.
That is, we are given two point sets: the assets $\points$ and the disk centers
$Y$, and a coverage function $\kappa$. Wanted are the radii of the disks
centered at $Y$ that meet the coverage requirement and minimize the sum of the
areas of the disks.  This is known as the non-uniform minimum-cost multi-cover
(MCMC) problem.
If $ \forall p \in \points, \kappa(p)=k $ it is referred to as the uniform MCMC.

The uniform MCMC, motivated by fault-tolerant sensor network design, was considered by Abu-Affash~et~al.~\cite{abu2011multi}, who presented an algorithm with a cost that is at most $23.02 + 63.95(\kappa_{\text{max}} - 1)$ times the cost of the optimal solution, where
$\kappa_{\text{max}}$ is the maximum coverage requirement.
The expected running time of the algorithm is $\bigO((n + m)\kappa_{\text{max}})$, where $n$ is the number of points of $\points$ and $m$  is the number of points of $Y$.

BarYehuda and Rawitz \cite{bar2013note} gave a $3^\alpha \kappa_{\text{max}}$-approximation algorithm for the non-uniform MCMC where the minimized cost is $\sum_{i=1}^{m} r_i^{\alpha}$.
The algorithm's expected running time   is polynomial.

Bhowmick~et~al.~\cite{bhowmick2013constant} also tackled the non-uniform MCMC problem.
The authors presented a polynomial-time algorithm achieving the first constant factor approximation for $\kappa(\points)>1$.
This work demonstrates that an approximation bound independent of $\kappa$ can be achieved.
The proposed algorithm recursively computes a $(\kappa - 1)$-cover and extends it to a $\kappa$-cover, relating the cost of this extension to the cost of a subset of disks referred to as the primary disks.
The algorithm's expected running time is polynomial.

Huang~et~al.~\cite{huang2021ptas,huang2024ptas} addressed the non-uniform MCMC problem.
The authors presented the first Polynomial Time Approximation Scheme (PTAS) for this problem for $\kappa(\points)>1$, providing a solution that can be made arbitrarily close to the optimal by choosing an appropriate $\epsilon$.
Their approach utilized techniques such as Balanced Recursive Realization and Bubble Charging, which allowed them to optimize the disks at a sub-disk level.
The expected running time of their $1+\epsilon$ approximation algorithm is $\bigO\left(n^{\bigO(1)} m^{\bigO(1/\epsilon) \bigO(d/\alpha)}\right)$, where the points are in $\mathbb{R}^d$ space and the minimized cost is $\sum_{i=1}^{m} r_i^{\alpha}$.
Instead of optimizing each disk as a whole, they show that it is possible to further approximate each disk with a set of sub-boxes and optimize them at the sub-disk level.
They first compute an approximate disk cover with minimum cost through dynamic programming, and then obtain the desired disk cover through a balanced recursive realization procedure.

Another line of closely related work examines placing the minimum number of unit disks to multi-cover a set of points.
As in our problem, the centers of these circles are arbitrary.
Unlike our problem, the radii are fixed and the number of unit disks is unbounded.
Gao~et~al.~\cite{gao2022fast} presented a 5-approximation algorithm with runtime $\bigO(n + \kappa_\text{max})$, and they introduced an improved 4-approximation algorithm with a higher time complexity of $\bigO(n^2)$. 
Filipov and Tomova tackled the problem of coverage with the minimum number of equal disks~\cite{Georgiev2023}.
They proposed a stochastic optimization algorithm of estimated time complexity $\bigO(n^2)$.

\section{Solving GMC: Lower Bounds}\label{sec:LB}

Any DGMC solution is a feasible solution for the GMC.
Thus, an optimal GMC solution automatically yields a lower bound for the variant with separation constraints.
This section contains two approaches for solving the GMC problem.

\subsection{\iteralg{}}\label{sec:approximation}

The heuristic determines good locations and radii of disks that collectively cover a set of points \( S \) with a specified coverage requirement \( \kappa (S)\).

In \textbf{Step 0}, the input points are shuffled to introduce randomness in the subsequent steps, thereby enhancing robustness across various input data configurations.

\textbf{Step 1}   uses $k$-means to cluster the set \( S \) into \( m \) clusters.  A minimum bounding disk is generated for the points assigned to each cluster.

In \textbf{Step 2}, the algorithm iteratively adjusts the clusters and their corresponding disk radii.
In iteration $i$, points not sufficiently covered by the current configuration are identified and assigned to their $i^\textrm{th}$ closest clusters where $i$ is the current coverage redundancy.
The clusters and their bounding disks are then updated accordingly until all points meet the coverage requirement \( \kappa(S) \).

\textbf{Step 3} addresses overcoverage by reducing the disk sizes where possible.
The algorithm identifies points that are covered more times than required and checks if removing a point from a cluster enables reducing the disk radius without causing undercoverage.
If this condition is satisfied, the point is removed from the cluster, and the disk is recomputed.

Through these steps, the algorithm ensures that the total area of the disks is locally optimized, resulting in an efficient covering of the point set \( S \) while meeting the coverage requirement.

\textbf{Special Case:} If \( m > |S| \), the algorithm proceeds by differentiating between two sub-cases:
 
If the total required coverage \( \sum_{p \in S} \kappa(p) \) is less than or equal to \( m \), each point $p$ in \( S \) is replicated $\kappa(p)$ times.
The algorithm allocates the required number of disks to directly cover each point, ensuring that the coverage requirement is satisfied.
This is done at zero cost meaning that all radii are zero.

If the total required coverage \( \sum \kappa(p) \) exceeds \( m \), the algorithm first assigns one disk to each point in \( S \).
The uncovered points are handled in an iterative process, where $k$-means clustering is applied to the uncovered points to place the remaining disks and the algorithm continues to adjust the disk radii until all points are sufficiently covered.

Computation time is dominated by step 2, which has time complexity  $\bigO( \kappa_{\max} n m)$.


\subsection{Integer Programming}\label{sec:IP}

An effective method for finding optimal solutions to NP-hard problems is Integer Programming~(IP).
Although solving an IP can take exponential time in the worst-case scenario, using meticulously designed mathematical models, specialized algorithm engineering, and existing IP solvers allows for solving considerably large instances to provable optimality.

To formulate the GMC as an IP, we need a discrete set of candidate sensor positions.
We discuss computing a (preferably small) sufficient set $C$ of candidate disks in \cref{section:candidate-disks}.
Given $C$, we can formulate the integer program in \cref{section:ip-formulation}.

\subsubsection{Computing the Candidate Set}
\label{section:candidate-disks}
Without separation constraints, there are three ways that a set of points $\points'\subseteq \points$ can be covered optimally (i.e., with minimum-area) by a disk.
\begin{enumerate}
    \item[a)] For $\points'=\{\point_1\}$, a disk centered at $\point_1$ with radius $0$ is optimal.
    \item[b)] For $\points'=\{\point_1,\point_2\}$, there is a unique disk with radius $\frac{\lVert \point_i- \point_j \rVert}{2}$ positioned at the midpoint of $\point_1, \point_2$ that is the minimum-area disk that contains both points.
    \item[c)] For $|\points'|\geq 3$, there is a minimum-area disk that has either (i) two points $\point_1, \point_2\in \points'$ on its boundary (which resembles case b) or (ii) at least three points $\point_1,\point_2,\point_3\in \points'$ on its boundary.
        This is because every other disk can be shrunk and translated until either (i) or (ii) is satisfied.
\end{enumerate}

The above properties give rise to a remarkably simple way of enumerating all necessary disks for the GMC.

We add a disk with radius $0$ and center $\point_i$ for all points $\point_i\in \points$.
Then for all pairs $\point_i, \point_j\in \points$, we compute the disk on the center between $\point_i,\point_j$ and radius $\frac{\lVert \point_i- \point_j \rVert}{2}$.
For all triples $\point_i, \point_j, \point_k\in \points$, we compute the unique disk that has $\point_i, \point_j$ and $\point_k$ on its boundary.
When the triangle between the points is obtuse, a disk in $C$ (defined by two of the points) already contains the third point and has a smaller area.
Thus, we only add a disk defined by three points if we encounter an acute triangle.

In total, this process yields $\bigO(n^3)$ possible disk positions.
For each disk, we need to find the points that are contained in the disk.
Using a $k$-$d$- or ball-tree one can find the set $\points'\subseteq \points$ of points that intersect a given disk in $\bigO(\sqrt{n}+|\points'|)$.
This yields a worst-case runtime of $\bigO(n^4)$ to enumerate all elements of $C$, but with better performance in practice.

\subsubsection{\gmcip{} Formulation}\label{section:ip-formulation}

For every disk $d_i$ in the candidate set $C$, we define integer variables $x_i$ that encodes how often each disk is used in the solution.
The constraints ensure that at most $m$ disks are placed and every point $\point_j\in \points$ is covered by at least $\kappa(\point_j)$ disks.
\begin{equation*}
    \begin{array}{ll@{}ll}
        \text{minimize}  & \displaystyle \pi \cdot \sum\limits_{d_i\in C} &r_{i}^2x_{i} &\\
        \text{subject to}& \displaystyle\sum_{d_i\in C}  &  x_{i} \leq m&\\
                         & \displaystyle\sum_{\substack{d_i\in C\\ \point_j \in d_i}}   &x_{i} \geq \kappa(\point_j),  & \forall \point_j \in \points\\
                         &&x_{i} \in \{0, \dots ,m\}, & \forall d_i \in C
    \end{array}
\end{equation*}

\section{Upper Bounds: Enforcing separation constraints}\label{sec:UB}

\begin{figure*}[t]
    \centering
    \includegraphics[width=.9\textwidth]{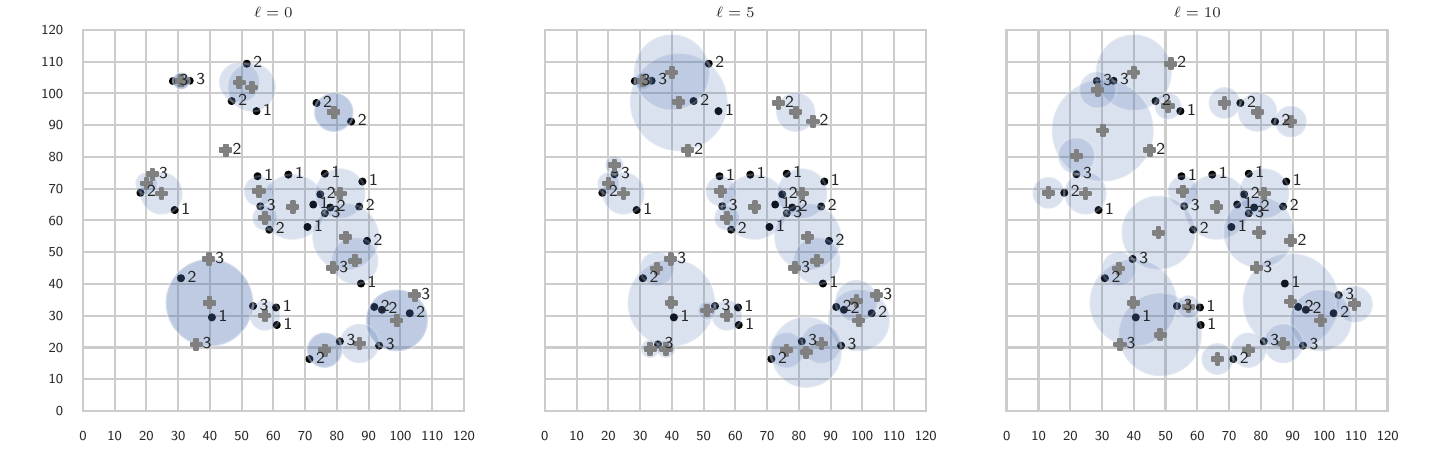}
    \caption{\label{fig:separation}
         Solutions from \largeFixM{} for the DGMC with different separation constraints; $\ell=0,5,10$, $n=40$ and $m=30$.
    }
\end{figure*}

Because the disk centers correspond to sensor locations,
we sometimes want to enforce a minimum separation between drones to avoid collisions or occlusions.
The DGMC is an extension of the GMC, for which the disks cannot be within $\ell$ distance; see \cref{fig:separation}.
One possible way of formulating the DGMC is introducing $3m$ continuous variables for all disks; $x,y$ coordinates, and radius $r$ (thereby not discretizing disks with a candidate set at all).
Convex SOCP (second-order cone programming) constraints can ensure coverage of the point set.
Disk separation, however, would require non-convex quadratic constraints, making finding an optimal solution significantly harder.

Therefore, we again work with a discretized candidate set $C$ and modify the Integer Programming formulation from~\cref{section:ip-formulation}.
However, unlike for the GMC, $C$ does not necessarily contain disks of an optimal solution.
It could even be that no selection of disks from $C$ provides any feasible solution to the DGMC.
Therefore, we modify $C$ to improve the quality of our solutions; see~\cref{sec:candidate-set-modification}.
Regarding quality certificates, we use the fact that the GMC is a lower bound to the DGMC.
Comparing this lower bound to any discretized DGMC solution allows us to evaluate the quality of the solution regarding the (non-discretized) DGMC.

\subsection{Introducing Separation Constraints}
We start by modifying the integer program from \cref{section:ip-formulation} to introduce a separation between two selected disks.
This is done by replacing the integer variables $x_i$ with binary variables $(x_i \in \{0,1\})$ to ensure that each disk can be selected at most once.
Given some distance $\ell$, we further add constraints that prevent two disks with distance $\leq \ell$ from being selected.
\begin{equation}
    x_i + x_j \leq 1 \hspace{3mm} \forall d_i, d_j \in C: \lVert y_i - y_j \rVert \leq \ell.
    \label{eq:separation}
\end{equation}

As the amount disks is $\bigO(n^3)$, this would yield  $\bigO(n^6)$ possible constraints.
In a real-world scenario many such constraints are unnecessary for finding an optimal solution.
Thus, we only add the violating constraints for two disks $d_i,d_j$ whenever the \gmcip{} finds a feasible solution.

Due to the separation requirement, we can further add a \emph{clique constraint} for each violating disk that excludes any two disks within a distance $< \frac{\ell}{2}$ from a feasible solution.
For some disk $d_i$ this clique constraint can be formulated as

\begin{equation}
    \sum_{\substack{d_j\in C\\ d(y_i, y_j) < \ell/2}} x_j \leq 1.
    \label{eq:separation-clique}
\end{equation}

\cref{eq:separation-clique} includes all pairwise separations from \cref{eq:separation}.
Thus we add constraints \cref{eq:separation} for violating disk pairs that are far apart, and \cref{eq:separation-clique} for the $< \frac{\ell}{2}$ neighborhood around each violating disk.
To limit the size of the resulting \dgmcip{}, \cref{eq:separation-clique} is added only for a disk $d_i$ if no clique was added for some other disk $d_j$ within a close neighborhood (i.e., $\lVert y_i - y_j \rVert < \frac{\ell}{2}$).

\subsection{Modifying the Candidate Set}\label{sec:candidate-set-modification}
The second idea is to extend the candidate set $C$ to include promising disks for coverage.
In the GMC solution, single outlier points $\point$ with $\kappa(\point)>1$ are often covered using $\kappa(\point)$ many drones that cover only $\point$.
This is no longer possible if we want to ensure any separation between disks.
For each point $\point$ with $\kappa(\point)>1$, we extend the candidate set $C$ by $\kappa(\point)$ small disks, that respect the separation constraints, i.e., we construct a regular $\kappa(\point)$-gon with side length $\ell$ centered around $\point$.

To speed up the solver, we can focus on \emph{small} disks.
To that end, we check for the largest disk $d_i$ used in the GMC solution and remove all other disks that have a radius that is greater than $\alpha\cdot r_i$.
The factor $\alpha$ compromises between the size of $C$ and solution quality.

While we cannot guarantee that the candidate set contains all disks from OPT, we experimentally confirm that found solutions are very close to lower bounds provided by the GMC IP, see~\cref{sec:Results}.

\section{Results} \label{sec:Results}

\begin{figure*}[t]
    \hfill
    \includegraphics[width=\columnwidth]{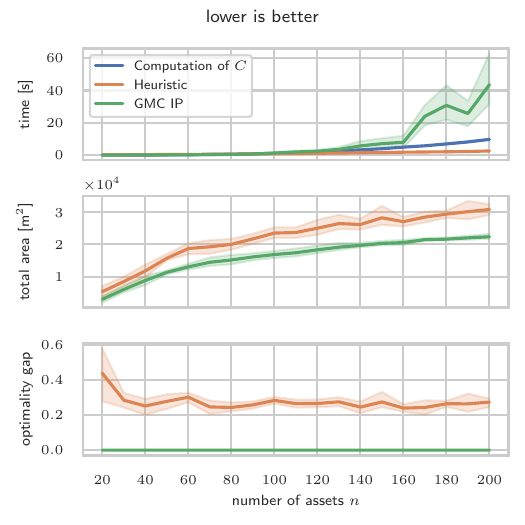}
    \hfill
    \includegraphics[width=\columnwidth]{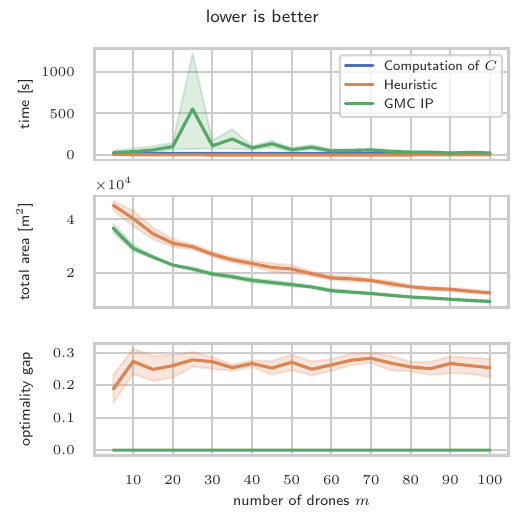}
    \hfill
    \caption{\label{fig:gmc}
        Comparison of runtime, total area, and optimality gap between the \gmcip{} solver and the heuristic. On all plots, lower is better. (Left) \smallFixM{}; fixed $m=20$ variable $n$ (Right) \fixN{}; fixed $n=250$ variable $m$.
    }
\end{figure*}

Experiments were carried out on a regular desktop workstation with an
AMD Ryzen 9 7900 ($12\times$\SI{3.7}{\giga\hertz}) CPU and \SI{88}{\giga\byte} of RAM.
Code and data are available\footnote{\url{https://gitlab.ibr.cs.tu-bs.de/alg/disc-covering}}.
Instances were generated uniformly in a $\SI{100}{\metre}\times \SI{100}{\metre}$ canvas.
Values of $\kappa(\point)$ for all points were sampled uniformly from $\{1,2,3\}$.
This yields the instances sets
\smallFixM{} (with $5$ instances for each size $20, 30, \dots, 200$ with a fixed value of $m=20$),
set \largeFixM{} (with $5$ instances for each size $30,40,\dots, 300$ with a fixed value of $m=30$),
and \fixN{} (with $5$ instances for each number of drones $5,10,\ldots, 100$ with a fixed value of $n=250$).

\subsection{GMC}

We compare the \iteralg{} and the integer program in terms of runtime and total area.
The experiments were run on benchmark sets \smallFixM{} and \fixN{}, which variate $n$ and $m$, respectively.

\subsubsection{Comparison of runtime cost}
\gmcip{} requires time to set up the solver, i.e.,
(i) computing the candidate set $C$ which takes $\bigO(n^4)$, see \cref{sec:IP},
and (ii) building the model which takes $\bigO(n|C|)$.
The solver is executed on the resulting integer program.

\cref{fig:gmc} shows that the runtime of the \gmcip{} solver is significantly higher than that of the \iteralg{}.
While computing the candidate set is challenging for larger instances, we do not observe the worst-case behavior in runtime.
For \fixN{} there exists a difficult interval for $m$ between $20$ and $35$ in which \gmcip{} needs significantly more time to obtain provable optimal solutions.

\subsubsection{Comparison of total area cost}

The lower row of \cref{fig:gmc} shows a comparison between \iteralg{} and \gmcip{} in terms of solution quality.
The plot shows the optimality gap that is the relative gap between the found solution versus the optimal solution ($(C_{\textrm{alg}}-C_{\textrm{ip}})/C_{\textrm{alg}}$).
For both fixed $m$ and $n$, the optimality gap remains stable at around $\SI{27.5}{\percent}$ in different instances, indicating that the performance is consistently close to optimal, regardless of the size of the problem.
The only exception being the case where $m \geq n$ in which the iterative algorithm gives slightly worse results.

\subsection{DGMC}

In \cref{sec:candidate-set-modification} we presented different candidate set strategies that are now compared in terms of solution quality and runtime.
Furthermore, we evaluated how different parameters affect the performance of the \dgmcip{}.
We ran these exploratory experiments on the benchmark set \smallFixM{}, as the workstations ran out of memory for larger instances.
After identifying the best parameters, we ran another experiment on the bigger benchmark set \largeFixM{}, i.e., for more points and more drones. 
For a fixed value of $\ell$, it is more reasonable to variate the number of points $n$, as more drones increase the difficulty of sparsifying the solution and lead to more infeasibilities.
For all the experiments, we chose a fixed $\ell=5$.

\subsubsection{Parameters of \dgmcip{}}

\begin{figure}[htb]
    \centering
    \includegraphics[width=1\columnwidth]{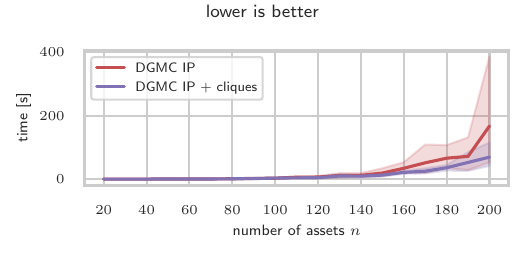}
    \caption{\label{fig:ipparam_clique_time}
    Influence of clique constraints on the runtime of the \dgmcip{} on \smallFixM{}; $m=20$ and $\ell=5$.}
\end{figure}

First, we investigated the impact of clique constraints and the set reduction on the runtime and solution quality the \dgmcip{}.
\cref{fig:ipparam_clique_time} shows the speedup for enabling the clique constraints.
As the time needed to solve smaller instances is low, the speedup is more significant for larger instances.
For the remainder of the experiments we invoked the clique constraints.

\begin{figure}[htb]
    \centering
    \includegraphics[width=1\columnwidth]{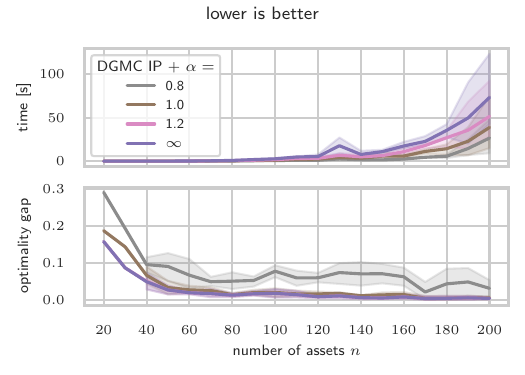}
    \caption{\label{fig:dgmc_small}
        Tradeoff between reducing the candidate set $C$ and the optimality gap for the \dgmcip{} on \smallFixM{};
    $m=20$ and $\ell=5$. For comparison, we only display instances that were feasible for all $\alpha$ values, removing $14$ of the $95$ instances.}
\end{figure}

\cref{fig:dgmc_small} shows the described tradeoff for the $\alpha$ parameter:
$\alpha$ controls the size of the largest dirk in the candidate set $C$.
The top row shows that reducing $C$ improves the solver times significantly.
At the same time, the bottom row shows how the solution quality decreases with a smaller set.
Setting $\alpha = 1.2$ provides excellent tradeoff between solution quality and runtime,
reducing runtime while almost maintaining the same solution quality as the original set, i.e., $\alpha = \infty$.
 
\subsubsection{Large benchmark \largeFixM{}}
First we ran the \gmcip{} on the benchmark set to obtain the lower bounds for the DGMC.
There is a single instance with $n=300$ that could not be solved within the memory limit;
this is ignored in the remainder.

Based on the results from \smallFixM{}, we ran the \dgmcip{} on the larger benchmark set \largeFixM{} with a time limit of \SI{900}{\second} for the solver.
We enabled clique constraints and set $\alpha = 1.2$.
Note that without reducing $C$ (i.e. with $\alpha = \infty$), we cannot reliably solve the larger instances, i.e., instances with more than $250$ points, as the integer program requires too much memory. 
Note that as we are comparing against the \gmcip{} solutions (without any seperation constraints), the gaps to an optimal solution of the DGMC are smaller than what can be seen in this evaluation.

\begin{figure}[htb]
    \centering
    \includegraphics[width=1\columnwidth]{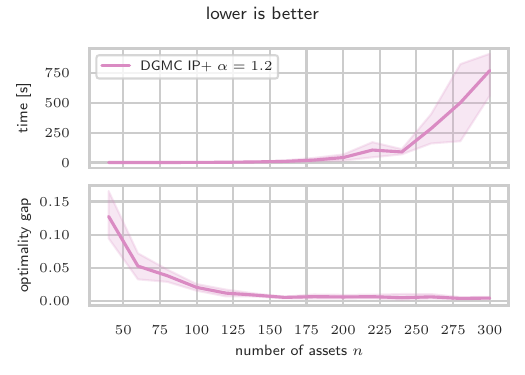}
    \caption{\label{fig:dgmc_large}
    Runtime and optimality gap of the \dgmcip{} on \largeFixM{}; $m=30$ and $\ell=5$.}
\end{figure}

\cref{fig:dgmc_large} shows that we can solve all instances close to provable optimality.
For smaller instances with $n\leq 100$ the optimality gaps are higher than $\SI{2}{\percent}$.
For larger instances, the \dgmcip{} solver was unable to find optimal solutions for the discretized DGMC (see~\cref{sec:UB}) and was terminated due to a timeout. Despite the early termination, \dgmcip{} found solutions with an optimality gap below \SI{0.7}{\percent} for all these instances. 

\section{Conclusions}

There are many avenues for future work.
These include an extension to covered assets in 3D, which is natural for many robot domains such as flying robots, space applications, or undersea sensor networks.
Calculating candidate centers is still possible, but more complicated~\cite{gartner1999fast}.
The high speed of the iterative approximation may be applicable for dynamic targets, or for adjusting the sensor positions and radii when a sensor is added or deleted.
A quadratic program for DGMC is suitable for small problems, and could be initialized with the \dgmcip{} solutions to speed computation.

\old{
\appendix
\newcommand{\diamP}{G}
\subsection{Iterative Algorithm's Pseudocode}
\input{Algorithms/heuristic}

\subsection{Quadratic Program}\label{sec:quadratic-program}
Let $[m]$ denote the set $\{1,\ldots,m\}$ and $\diamP$ denote the diameter of the point set $\points$.
For a point $\point$, let $x(\point)$ and $y(\point)$ denote the $x$- and $y$-coordinate of $\point$, respectively.
The following quadratic program can be used to solve the DGMC to provable optimality.
\begin{equation*}
    \begin{array}{lll}
        \text{min}  & \displaystyle
        \pi \cdot \sum\limits_{i\in [m]} r_{i}^2 &\\
        \text{s.t.}& \displaystyle
        (x_{i} - x(\point_j))^2 + \\
                   & (y_{i} - y(\point_j))^2 - \\
                   & \left((1- c_{j,i}) \cdot \sqrt{2}\diamP\right)^2 \leq r_{i}^2& \forall \point_j,i \in \points\times [m]
                   \\  \displaystyle
        \sum_{i\in [m]}   & c_{j, i} \leq \kappa(\point_j)& \forall \point_j \in \points
                       \\ & \displaystyle
        (x_{i}^2 - x_{k}^2) +  (y_{i}^2 - x_{k}^2) \geq \ell^2 & \forall i,k \in [m] \times [m]
        \\
                                                               &x_{i} \in [-\diamP, \diamP], & \forall i\in[m]\\
                                                               &y_{i} \in [-\diamP, \diamP], & \forall i\in[m]\\
                                                               &r_{i} \in [-\diamP, \diamP], & \forall i\in[m]\\
                                                               &c_{j,i} \in \{0,1\} & \forall \point_j,i \in \points\times [m]
    \end{array}
\end{equation*}
The variables $x, y$ and $r$ are continues variables that represent the center and radius of a disk.
The Boolean variable $c_{j,i}$ decides if the disk $i$ covers the point $\point_j$.

The objective function minimizes the total area of the disks.
The first constraint ensures if $c_{j,i}$ is true, i.e., the dirk $i$ is suppose to cover the point $\point_j$, the radius of the disk is large enough to do so, otherwise this constraint is not active.
The second constraint ensures that each point $\point_j$ is covered by at most $\kappa(\point_j)$ disks.
The last constraint ensures the separation between the disks.

\subsection{Computing the candidate set}\label{sec:identiying-acute-triangles}

To minimize the size of the candidate set $C$, we can ignore disks that are defined by three points that form an obtuse triangle, see~\cref{section:candidate-disks}.
Let $\Delta$ be the triangle defined by three points $\point_i,\point_j,\point_k$.
The three side lengths of $\Delta$ are denoted as $a,b,c$ (w.l.o.g. $a\leq b \leq c$).

The triangle $\Delta$ is obtuse iff $a^2 + b^2 \leq c^2$.
In the case where the triangle is acute, we compute the center of the disk $d$ by intersecting the perpendicular bisectors of the sides.
To find its coordinates, we solve the linear system derived from the equations of these perpendicular bisectors.
The radius is $r = \frac{abc}{4A(\Delta)}$
where $A(\Delta)$ is the area of $\Delta$.
Using Heron's formula $
A = \sqrt{s(s-a)(s-b)(s-c)}$
where \( s \) is the semi-perimeter $s = \frac{a + b + c}{2}$ we obtain
\begin{equation*}
    r = \frac{abc}{\sqrt{(a + b + c)(b + c - a)(c + a - b)(a + b - c)}}.
\end{equation*}
\subsection{Problem size depends on point distribution}
The number of candidate disks $|C|$ is dominated by the number of acute triangles, which depends on the distribution of the set of points.
Eric Langford solved the problem ``Given three points at random in an $1\times L$ rectangle, what is the probability $P(L)$ that the triangle thus formed is obtuse?''~\cite{langford1970problem}.
$P(L)$ has a minimum for the case $L=1$, i.e. when the rectangle is a square.
\begin{align*}
    P(1) = 97/150+\pi/40 \approx 0.725206
\end{align*}
If instead the points in $\points$ are sampled uniformly from the unit disk, Woolhouse~\cite{woolhouse1886solution} obtained
\begin{align*}
    P_\bigcirc=1-\left({{\frac{4}{\pi^2}}-{\frac{1}{8}}}\right) \approx 0.719715 \, .
\end{align*}
}

\clearpage
\IEEEtriggeratref{9}
\bibliographystyle{IEEEtranS}
\bibliography{IEEEabrv,biblio.bib}
\end{document}